\documentclass[10pt,twocolumn,letterpaper]{article}

\usepackage[pagenumbers]{cvpr} 

\usepackage{graphicx}
\usepackage{amsmath}
\usepackage{amssymb}
\usepackage{booktabs}
\usepackage[ruled]{algorithm2e}
\usepackage{amsmath}
\usepackage{stfloats}
\usepackage{multirow}
\usepackage{float}
\usepackage{stfloats}
\usepackage{indentfirst}

\usepackage[pagebackref,breaklinks,colorlinks]{hyperref}

\usepackage[capitalize]{cleveref}

\crefname{section}{Sec.}{Secs.}
\Crefname{section}{Section}{Sections}
\Crefname{table}{Table}{Tables}
\crefname{table}{Tab.}{Tabs.}

\def\cvprPaperID{XX} 
\def\confName{CVPR}
\def\confYear{2022}

\begin{document}

\title{Eliminate Deviation with Deviation for Data Augmentation and a General Multi-modal Data Learning Method}

\newcommand*{\affaddr}[1]{#1} 
\newcommand*{\affmark}[1][*]{\textsuperscript{#1}}
\newcommand*{\email}[1]{\texttt{#1}}
\author{%
	Yunpeng Gong\affmark[]\affmark[]\qquad Liqing Huang\affmark[]\qquad Lifei Chen\affmark[] \\
	\affaddr{\affmark[]College of Computer and Cyber Security, Fujian Normal University, P. R. China}\qquad
	\affaddr{\affmark[]}\\
	\email{\tt\small \affmark[]fmonkey625@gmail.com}\qquad
	\email{\tt\small \affmark[]\{lqhuang,clfei\}@fjnu.edu.cn}
}
\maketitle

\begin{abstract}
One of the challenges of computer vision is that it needs to adapt to color deviations in changeable environments. Therefore, minimizing the adverse effects of color deviation on the prediction is one of the main goals of vision task. Current solutions focus on using generative models to augment training data to enhance the invariance of input variation. However, such methods often introduce new noise, which limits the gain from generated data. To this end, this paper proposes a strategy eliminate deviation with deviation, which is named Random Color Dropout (RCD). Our hypothesis is that if there are color deviation between the query image and the gallery image, the retrieval results of some examples will be better after ignoring the color information. Specifically, this strategy balances the weights between color features and color-independent features in the neural network by dropouting partial color information in the training data, so as to overcome the effect of color devitaion. The proposed RCD can be combined with various existing ReID models without changing the learning strategy, and can be applied to other computer vision fields, such as object detection. Experiments on several ReID baselines and three common large-scale datasets such as Market1501, DukeMTMC, and MSMT17 have verified the effectiveness of this method. Experiments on Cross-domain tests have shown that this strategy is significant eliminating the domain gap. Furthermore, in order to understand the working mechanism of RCD, we analyzed the effectiveness of this strategy from the perspective of classification, which reveals that it may be better to utilize many instead of all of color information in visual tasks with strong domain variations.
\end{abstract}

\section{Introduction}
\label{sec:intro}

\begin{figure}[]
	\centering
	\includegraphics[width=1\linewidth]{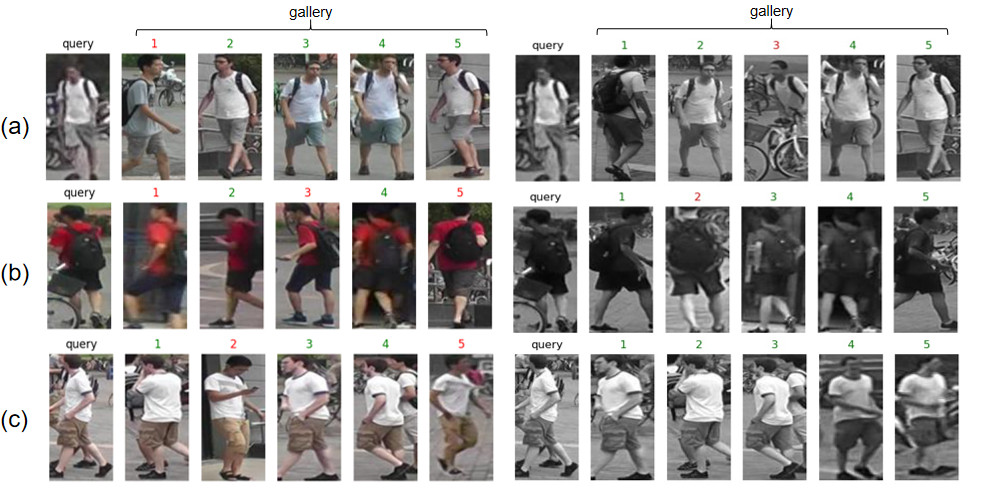}
	\caption{The retrieval results of the model trained with visible (RGB) image and the model trained with grayscale image on Market1501\cite{market1501} are displayed. It shows that the color deviation between the query image and gallery image will affect the retrieval results, and the retrieval results of some samples will be better after ignoring the color information. The numbers on the images indicate the rank of similarity in the retrieval results, the red and green numbers denote the wrong and correct results, respectively.}
	\label{fig:onecol}
\end{figure}
Person re-identification (ReID) is to match the same person across diferent cameras and scenes\cite{survey,market1501,baseline,Li_2021_CVPR}. This technology have been widely applied to video surveillance\cite{Hou_2021_CVPR,Tian_2021_CVPR,Liu_2021_CVPR}, image retrieval\cite{sketch-criminal,sketch-based}, criminal investigation\cite{sketch-criminal}, target tracking\cite{Beyer_2017_CVPR_Workshops} and others. The challenge of this task is that images captured by different cameras often contain significant intra-class variation caused by variations in viewpoint, human pose changes, occlusions, and color deviation under variable camera conditions, etc. As a result, the appearance of the same pedestrian image with great changes, making intra-class (the same pedestrian) metric distance larger than inter-class (different pedestrians). ReID usually combines representation learning\cite{zheng2017discriminatively,matsukawa2016person,fan2019spherereid,lin2019improving,zheng2016person,yao2019deep} with metric learning\cite{wang2019ranked,varior2016gated,varior2016siamese,schroff2015facenet,liu2017end,cheng2016person,hermans2017defense}, and combines classification loss\cite{yao2019deep,fan2019spherereid,zheng2017discriminatively} with triplet loss\cite{hermans2017defense,cheng2016person} in the training stage to optimize the neural network. In the inference stage, it is only necessary to use Cosine distance or Euclidean distance to measure the similarity between the query image and the gallery image, and then rank the gallery images according to the similarity, and finally use the re-ranking technique\cite{re_ranking} to further refine the search results.

The complexity of the inherent challenge of ReID means that its demand for data has the same complexity, and the complex data demand is difficult to meet and balance by the training set, which is also accompanied by the potential problem that the model overfits the only training data and lacks robustness. The dataset is hard to cover different camera environments and all their variations at different times, so the trained models tend to overfit the given training set and lack robustness to additional scenarios. It is no doubt that color features are important discriminative features, but color features instead limit the model to make correct predictions in some cases. For example, because white and gray, black and dark blue, and brown and yellow are similar under some lighting conditions, it is difficult for the model to make correct predictions for negative samples that are similar to target after overfitting the color deviation variations. As shown in (a) to (c) in Figure 1, the prediction results made by the model trained with grayscale images in this case are better after discarding the color bias.

Realistically, color deviations cause domain gaps exist both between datasets and within datasets\cite{zheng2019joint,wei2018person}. These color biases are practically inexhaustible. Instead of generating a variety of data to let the model "see" these variations (especially intra-class variations)\cite{zheng2019joint} during training to enhance the robustness to input variations, it is better to balance the weight between color features and other important discriminant features implicitly. Aiming at the inherent color deviation problem that the images obtained under different shooting conditions, this paper proposes a strategy to eliminate deviation with deviation based on the assumption that the retrieval results of some samples will be better when discarding color information, which is named random color dropout (RCD). RCD balances the weight between color features and other important discriminant features in neural network by discarding part of the color information in the training data, so as to overcome the influence of color deviation. This strategy exists in various forms. For example, color deviation can be overcome with biased grayscale information or sketch information (or contour information). Taking grayscale as an example, it can be randomly selected a rectangular area in the RGB image and replace its pixels with the same rectangular area in the corresponding grayscale image , thus it generates a training image with different areas of biases during ReID model training. Compared with existing methods based on generative adversarial networks (GANs)\cite{goodfellow2014generative,zheng2019joint,wei2018person,zhong2018camera,deng2018image,liu2018pose,qian2018pose}, the proposed method is more lightweight and effective because it not only does not introduce new noise but also saves a large amount of computational resources. At the same time, this strategy enables the model to naturally have cross-modal retrieval\cite{sketch-criminal,sketch-based,zhu2020hetero,huang2022modality,ye2018visible,li2020infrared} capabilities. For example, when taking the contour information as the intermediary to overcome the color deviation, the cross-modal retrieval between sketch and RGB visible image can be realized.

In addition, this paper analyzes the relationship between RCD and the generalization ability of neural networks from the perspective of classification, and reveals the intrinsic reasons that networks trained with RCD may outperform ordinary networks. Experiments show that the proposed method not only increases the robustness of the model to color deviation but also bridges the domain gap between different datasets, which has significant advantages over the existing state-of-the-art method. Taking the grayscale as an example, the RCD strategy proposed in this paper includes global grayscale transformation, local grayscale transformation, and a combination of these two. The method has the following advantages:

It is a lightweight approach which does not require any additional parameter learning or memory consumption. It can be combined with various CNN models without changing the learning strategy. It is a complementary approach to existing data augmentation. 

The main contributions of this paper are summarized as follows:

$\bullet$ This paper proposes a learning strategy which against color deviation with information deviation, which decreases the overfitting and increases generalization ability of the model.

$\bullet$ A simple and effective cross-modal retrieval method is proposed, which does not need complex network design.

$\bullet$ This paper proves that the network trained with RCD may be better than the ordinary network from the perspective of classification.

$\bullet$ The strategy proposed in this paper is proved to be effective in improving ReID performance through extensive experiments and analysis. The effectiveness of the proposed method is verified on several baselines and representative datasets.

This work was previously published as a preprint on Arxiv and extended on the basis of it, including related demonstration and cross-modal retrieval.

\begin{figure*}[htbp] 
	\setlength{\abovecaptionskip}{0.1cm}
	\setlength{\belowcaptionskip}{-0.4cm}   
	\centering
	\includegraphics[width=0.8\linewidth]{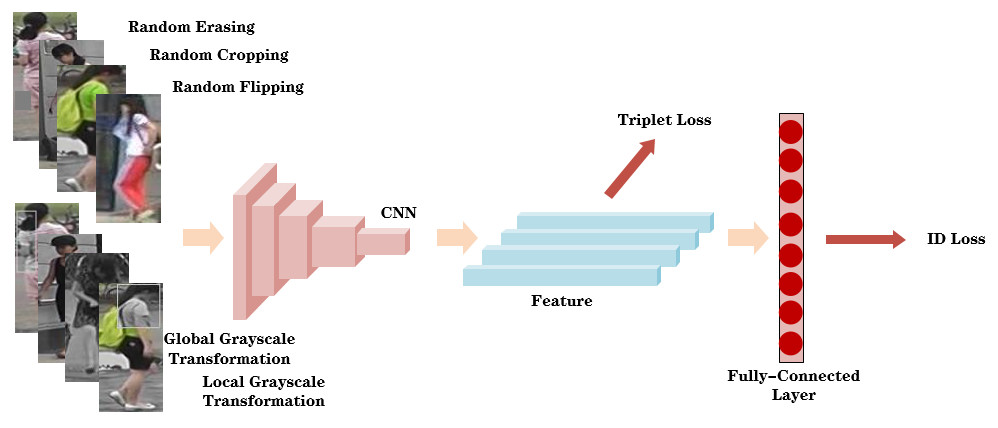}
	\caption{Framework diagram of our Random Color Dropout (RCD): The application of global grayscale transformation and local grayscale transformation in the framework.}
\end{figure*}
\section{Related Work}

The complexity of the inherent challenge of ReID means that its demand for data has the same complexity, and the failure to fully meet the complex demand of the data is the source of the problem of overfitting and insufficient generalization of the model to the training data. Improving generalization ability is the focus of research in convolutional neural networks (CNNs). Therefore, data augmentation is effective in improving the generalization ability of the model.
\subsection{Classic Data Augmentation}
Many data augmentation\cite{krizhevsky2012imagenet} methods have been proposed, such as random cropping\cite{krizhevsky2012imagenet}, flipping\cite{simonyan2014very}, which are well known to play an important role in classification, detection  and ReID. CutMix\cite{yun2019cutmix} replaces one patch
of an image with a patch from another image. Random erasing or cutout\cite{zhong2020random,devries2017improved} adds noise block to the image to regularize the network, while it helps to solve the occlusion problem in the ReID. The above methods are regarded as indispensable methods, and they are applied to various baselines\cite{luo2019bag,he2020fastreid,zheng2020vehiclenet,zheng2018discriminatively}. These techniques have been proved to be effective in improving the prediction accuracy, and they are complementary to each other\cite{luo2019bag}.

In solving the problem of color deviation, the early work\cite{li2014deepreid} used the filter and the maximum grouping layer to learn the illumination transformation, divided the pedestrian image into more small pieces to calculate the similarity, and uniformly handled the problems of misalignment, occlusion and illumination variation under the deep neural network; \cite{liao2015person} performed pre-processing before feature extraction and used multiscale Retinex algorithm to enhance the color information of light shaded regions to improve the color changes caused by lighting condition changes. 

With the increasing maturity of GANs, GANs-based approaches for data augmentation have become an active research field.
\subsection{Data Augmentation Based on GANs}	
The goal of these methods is to mitigate the effect of color deviation or human-pose variation, and to improve the robustness of the model by learning the invariant features from the variation of the input. The appearance details and the emphases generated by different GANs-based methods are also different, but their goal is all to compensate for the difference between the source and target domains. For example, CamStyle [52] generates new data for transferring different camera styles to learn invariant features between different cameras to increase the robustness of the model to camera style changes; CycleGAN\cite{zhu2017unpaired} was applied in\cite{deng2018image,zhong2019invariance} to transfer pedestrian image styles from one dataset to another; StarGAN\cite{choi2018stargan} was used by\cite{zhong2018generalizing} to generate pedestrian images with different camera styles. Wei et al.\cite{wei2018person} proposed PTGAN to achieve pedestrian image transfer across different ReID datasets. It uses semantic segmentation to extract foreground masks to assist style transfer, and converts the background into the desired style of the dataset while keeping the foreground unchanged. different from global style transfer, DGNet\cite{zheng2019joint} utilizes GANs to transfer clothing among different pedestrians by manipulating appearance and structural details to generate more diverse data to reduce the impact of color changes on the model, which effectively improves the generalization ability of the model. In addition,\cite{bak2018domain} uses 3D engine and environment rendering technology to build a virtual pedestrian data set with multiple lighting conditions, which is combined with other large real data sets to jointly pre-train a model.

The method proposed in this paper has been partially validated in other works. \cite{Gong_2022_CVPR} proved that the lack of robustness to color deviation is one of the main reasons why the model is vulnerable to adversarial metric attacks\cite{bouniot2020vulnerability,bai2019metric,wang2020transferable}, and enhanced the model's adversarial defense using the method proposed in this paper; It is adopted in the new baseline proposed in\cite{ni2021flipreid,chen2021benchmarks} to help increase the generalization of the model. In addition, \cite{ryu2021detection} showed that the method proposed in this paper is also suitable for object detection.

\section{Proposed Methods}
\label{sec:}
The RCD strategy proposed in this paper includes global transformation, local transformation, and a combination of the two. Taking grayscale as an example, it includes global grayscale transformation, local grayscale transformation, and combinations of the two. At the end of this subsection, we give the corresponding analysis of the proposed method. The framework of this method is showed in Figure $\color{red}2$.

\subsection{Global Grayscale Transformation}
In the data loading, it randomly samples K identities and M images of per person to constitute a training batch which size equals to $B=K\times M$. The set is denoted as $x^{v}=\{x_i^{v}|i=1,2,...,M\times K\}$, where $x_i^{v}=\{x_i^{v}|y_i \}$ represents the i-th sample image of the training batch, and $y_i$ represents the class label of the pedestrian.

Taking the grayscale as an example, this method randomly performs global grayscale transformation on the training batch with a probability, and then inputs into the model for training. This process can be defined as:
\begin{equation}
	I_{g} = t(R,G,B)
\end{equation}
where $t(\bullet)$ is the grayscale image conversion function, which is implemented by performing pixel-by-pixel accumulation calculations on the R, G, and B channels of the original visible RGB image; y is the label of the sample, the converted grayscale image label of $x^g$ are the same as the original ones:
\begin{equation}
	(x^{g}|y) = (x^v|y)
\end{equation}
the procedure of LGT is shown in Algorithm.$\color{red}1$.

\begin{algorithm}[t]
	\SetAlgoLined
	\SetKwInOut{Input}{Input}
	\SetKwInOut{Output}{Output}
	\SetKwInput{Initialization}{Initialization}
	\caption{Global Graycale Transformation}\label{algorithm 1}
	\Input{Input image $I$; \\
		Graycale transformation probability $p$; \\
		.}
	\Output{Grayscale images $I^{\ast}$.}
	\Initialization{$p_1 \leftarrow $ Rand (0, 1).} 
	
	\eIf{$p_1 \geq p$}{
		$I^{\ast} \leftarrow I$; \\
		\Return{$I^{\ast}$}.
	}{
		$I^{\ast} \leftarrow $ t($I$);\\
		\Return{$I^{\ast}$}.
	  }
\end{algorithm}

\subsection{Local Grayscale Transformation}
In addition to transforming the data globally, we also consider transforming the data locally so that the model adapt better to the significantly varying bias due to color dropout from the local variation.

The local grayscale transformation (LGT) for each visible image $x^v$ can be achieved by the following equations:

\begin{equation}
	x^g = t(x^v),
\end{equation}

\begin{equation}
	rect = RandPosition(x^v),
\end{equation}

\begin{equation}
	x^{lg} = LGT(x^v,x^g,rect)
\end{equation}
and
\begin{equation}
	(x^{lg}|y) = (x^v|y)
\end{equation}
where $x^g$ is the grayscale images, and $t(\bullet)$ is the grayscale transformation funtion; $RandPosition(\bullet)$ is used to generate a random rectangle in the image, and the function of $LGT(\bullet)$ is to give the pixels in the rectangle corresponding to the $x^g$ image to the $x^v$ image; $x^{lg}$ is the sample after local grayscale transformation, and $y$ is the label of the transformed image.


\begin{algorithm}[t]
	\SetAlgoLined
	\SetKwInOut{Input}{Input}
	\SetKwInOut{Output}{Output}
	\SetKwInput{Initialization}{Initialization}
	\caption{Local Graycale Transformation}\label{algorithm 2}
	\Input{Input image $I$; \\
		Image size $W$ and $H$; \\
		Area of image $S$; \\
		Transformation probability $p$; \\
		Area ratio range $s_l$ and $s_h$; \\      
		Aspect ratio range $r_1$ and $r_2$.}
	\Output{Transformed image $I^{\ast}$.}
	\Initialization{$p_1 \leftarrow $ Rand (0, 1).} 
	
	\eIf{$p_1 \geq p$}{
		$I^{\ast} \leftarrow I$; \\
		\Return{$I^{\ast}$}.
	}{
		\While {True}{
			$S_t\leftarrow $ Rand $(s_l, s_h)$$\times S$;\\
			$r_t \leftarrow $ Rand $(r_1, r_2)$;\\
			$H_t \leftarrow \sqrt{S_t \times r_t}$,~ $W_t \leftarrow \sqrt{\frac{S_t}{r_t}}$;\\
			$x_t \leftarrow $ Rand $(0, W)$,~ $y_t \leftarrow $ Rand $(0, H)$;\\
			\If{$x_t + W_t \le W$ and $y_t + H_t \le H$}{
				$Position \leftarrow (x_t, y_t, x_t+W_t, y_t+H_t)$;\\
				$I(Position) \leftarrow $ $t(Position)$;\\
				$I^{\ast} \leftarrow I$;\\
				\Return{$I^{\ast}$}.
			}
		}
	}
\end{algorithm}
In the process of model training, we conduct LGT randomly transformation on the training batch with a probability. For an image $I$ in a batch, denote the probability of it undergoing LGT be $p_r$, and the probability of it being kept unchanged be $1-p_r$. In this process, it randomly selects a rectangular region in the image and replaces it with the pixels of the same rectangular region in the corresponding grayscale image. Thus, training images which include regions with different levels of grayscale are generated. Among them, $s_l$ and $s_h$ are the minimum and maximum values of the ratio of the image to the randomly generated rectangle area, and the $S_t$ of the rectangle area limited between the minimum and maximum ratio is obtained by $S_t$ ← $Rand (s_l ,s_h ) \times S$, $r_t$ is a coefficient used to determine the shape of the rectangle. It is limited to the interval ($r_1$, $r_2$ ). $x_t$ and $y_t$ are randomly generated by coordinates of the upper left corner of the rectangle. If the coordinates of the rectangle exceed the scope of the image, the area and position coordinates of the rectangle are re-determined. When a rectangle that meets the above requirements is found, the pixel values of the selected region are replaced by the corresponding rectangular region on the grayscale image converted from RGB image. As a result, training images which include regions with different levels of grayscale are generated, and the object structure is not damaged. The procedure of LGT is shown in Algorithm.$\color{red}2$.

\subsection{Loss function}
ReD usually combines classification loss and triplet loss to train the model\cite{stong_baseline,FastReID}. Here, we use $x_i^{v}$ to denote the $i$-th RGB image in a training batch, and $x_i^{g}$ to denote the image obtained after GGT or LGT conversion. Thenthe features of $x_i^{v}$ and  $x_i^{g}$ can be expressed as:
\begin{equation}
	\left\{
	\begin{array}{ll}
		f_i^{v}=f(x_i^{v})\\
		f_i^{g}=f(f_i^{g})
	\end{array}
	\right.
\end{equation}
The Euclidean distance between two samples $x_i^{g}$ and $x_j^{g}$ is denoted as $D(x_i^{g},x_j^{g})$, where The subscript $\{ i, j \}$ denotes the image index in the training batch. Formally, let $x_i^{g}$ be the anchor sample, the triple $\{ x_i^{g},x_j^{g}, x_k^{g}\}$ is selected in the following way:
\begin{equation}
	P_{i,j}^{g} = \max_{\forall y_i=y_j}D(x_i^{g},x_j^{g})
\end{equation}
\begin{equation}
	N_{i,k}^{g} = \min_{\forall y_i\ne y_j}D(x_i^{g},x_k^{g})
\end{equation}
For each anchor point $x_i^{g}$, the above strategy selects the positive sample pair with the same pedestrian class label and the farthest the most distant positive sample pair with the same pedestrian category label and the nearest negative sample pairs, forming a triplet $\{ x_i^{g},x_j^{g^{+}}, x_k^{g^{-}} \}$ for mining grayscale information. In general, the use of The boundary parameter $\varepsilon$ is used to control the spacing of the positive and negative sample pairs. In summary, we can define the following triadic loss for training:
\begin{equation}
	L_{g} = \frac{1}{n}\sum_{i=1}^{n}\max[\varepsilon+D(x_i^{g},x_j^{g^{+}})+D(x_i^{g},x_k^{g^{-}})]
\end{equation}
In addition, $x^v$ and $x^g$ are trained using a shared identity classifier $\phi$. The predicted probability of identity labels $y_i$ is define as $p(y_i|x_i^{g};\phi)$. The ID loss is represented as follows:
\begin{equation}
	L_{ide} = -\frac{1}{n}\sum_{i=1}^{n}log(p(y_i|x_i^{g};\phi)) 
\end{equation}
Therefore, the overall loss during random grayscale transformation is:
\begin{equation}
	L_{total}=L_{g}+L_{ide} 
\end{equation}

\subsection{Analysis of Random Color Dropping Policy}
Here suppose there are $m$ instances, the expected output, i.e. $D = [d_1, d_2, …, d_m]^T$ where $d_j$ denotes the expected output on the $j$-th instance, and the actual output of the $i$-th component neural network, i.e. $F_i=[f_{i1}, f_{i2}, …, f_{im}]^T$ where $f_{ij}$ denotes the actual output of the $i$-th component network on the $j$-th instance. $D$ and $F_i$ satisfy that $d_j$ $\in \{-1, +1\} (j = 1, 2, …, m)$ and $f_{ij}$$\in\{-1, +1 \} (i = 1, 2, …, N; j = 1, 2, …, m ) $ respectively. It is obvious that if the actual output of the $i$-th component network on the $j$-th instance is correct according to the expected output then $f_{ij}d_j = +1$, otherwise $f_{ij}d_j = -1$. Thus the generalization error of the $i$-th component neural network on those $m$ instances is:

\begin{equation}
	E_i = \frac{1}{m}\sum_{j=1}^m {Error(f_{ij}d_j)}
\end{equation}
where $Error(x)$ is a function defined as:
\begin{equation}
	Error(x)=\left\{
	\begin{array}{ll}
		1, \quad\quad if \quad x=-1\\
		0.5, \quad if \quad x=0\\
		0, \quad\quad if \quad x=1
	\end{array}
	\right.
\end{equation}

Here we introduce a vector $Sum = [Sum_1, Sum_2, …, Sum_m]^T$ where $Sum_j$ denotes the sum of the actual output of all the component neural networks on the $j$-th instance, i.e.

\begin{equation}
	Sum_j = \sum_{i=1}^N f_{ij}
\end{equation}

Then the output of the neural network ensemble on the j-th instance is:

\begin{equation}
	\hat{f_j} = Sgn(Sum_j)
\end{equation}

where $Sgn(x)$ is a function defined as:

\begin{equation}
	Sgn(x)=\left\{
	\begin{array}{ll}
		1, \quad\quad if \quad x>0\\
		0, \quad\quad if \quad x=0\\
		-1, \quad if \quad x<0
	\end{array}
	\right.
\end{equation}

It is obvious that $\hat{f}_j\in \{-1, 0, +1 \} (j = 1, 2, …, m)$  . If the actual output of the ensemble on the $j$-th instance is correct according to the expected output then $ \hat{f_j}d_j = +1$; if it is wrong then $ \hat{f}_jd_j = -1$; otherwise $ \hat{f}_jd_j = 0$, which means that there is a tie on the j-th instance, e.g. three component networks vote for +1 while other three networks vote for -1. Thus the generalization error of the ensemble is:

\begin{equation}
	\hat{E} = \frac{1}{m}\sum_{j=1}^m {Error(\hat{f_{j}}d_j)}
\end{equation}

Here suppose that the k-th component neural network is trained using grayscale images. Then the output of the new set on the $j$-th instance is:
\begin{equation}
	\hat{f_j^{'}} = Sgn(Sum_{j(j \neq k)} - f_{kj})
\end{equation}
and the generalization error of the new ensemble is:
\begin{equation}
	\hat{E^{'}} = \frac{1}{m}\sum_{j=1}^m {Error(\hat{f_{j}^{'}}d_j)}
\end{equation}
It is assumed that a certain number of networks with deviations will not affect the performance of the overall neural network, and the retrieval results of some examples will be better after ignoring the color information.
\begin{equation}
	Error(\hat{f_j^{'}}d_j) \leqslant Error(\hat{f_j}d_j)
\end{equation}
From Eq.$\color{red}(18)$ and Eq.$\color{red}(20)$ we can derive that if Eq.$\color{red}(21)$ is satisfied then $\hat{E}$ is not smaller than $\hat{E^{'}}$, which means that the ensemble including k-th component neural network which is trained using grayscale images is better than the one no including:
\begin{equation}
\begin{split}
	\sum_{j=1}^m\{Error(Sgn(Sum_j)d_j)- \\Error(Sgn(Sum_{j(j \neq k)+f{kj}})d_j) \} \geq 0
\end{split}
\end{equation}

\section{Comparison and Analysis}
\begin{figure}[t]
	\setlength{\abovecaptionskip}{0.1cm}
	\setlength{\belowcaptionskip}{-0.4cm}   
	\centering
	\includegraphics[width=1\linewidth]{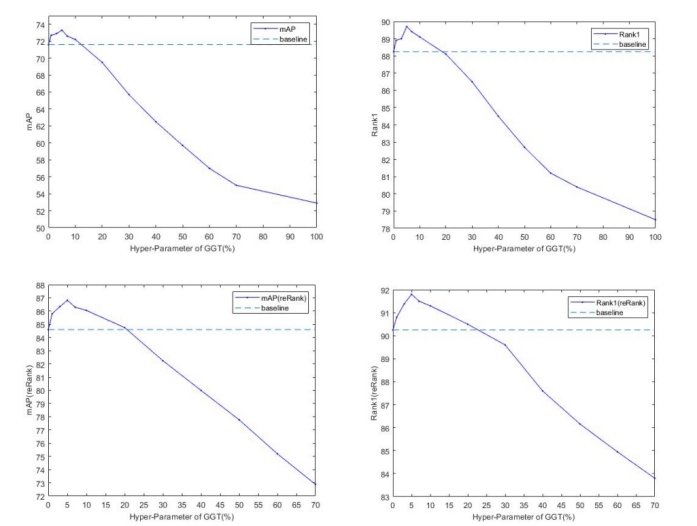}
	\caption{Performance of GGT under different hyperparameters on Market1501.}
\end{figure}
\subsection{Datasets and Evaluation criteria}
\textbf{Datasets}. Market-1501~\cite{market1501} includes 1,501 pedestrians captured by six cameras (five HD cameras and one low-definition camera). DukeMTMC~\cite{duke} is a large-scale multi-target, multi-camera tracking dataset, a HD video dataset recorded by 8 synchronous cameras, with more than 2,700 individual pedestrians. The above two datasets are widely used in ReID studies. MSMT17\cite{wei2018person}, created in winter, was presented in 2018 as a new, larger dataset closer to real-life scenes, containing a total of 4,101 individuals and covering multiple scenes and time periods.

These three datasets are currently the largest datasets of ReID, and they are also the most representative because they collectively contain multi-season, multi-time, HD, and low-definition cameras with rich scenes and backgrounds as well as complex lighting variations.

Sketch ReID dataset\cite{sketch-criminal} contains 200 persons, each of which has one sketch and two photos. Photos of each person were captured during daytime by two cross-view cameras. It cropped the raw images (or video frames) manually to make sure that every photo contains the one specific person. It have a total of 5 artists to draw all persons’sketches and every artist has his own painting style.

\textbf{Evaluation criteria}. Following existing works~\cite{market1501}, Rank-k precision and mean Average Precision (mAP) are adapted as evaluation metrics. Rank-1 denotes the average accuracy of the first return result corresponding to each query image. mAP denotes the mean of average accuracy, the query results are sorted according to the similarity, the closer the correct result is to the top of the list, the higher the score.

\begin{figure}[t]
	\setlength{\abovecaptionskip}{0.1cm}
	\setlength{\belowcaptionskip}{-0.4cm}   
	\centering
	\includegraphics[width=1\linewidth]{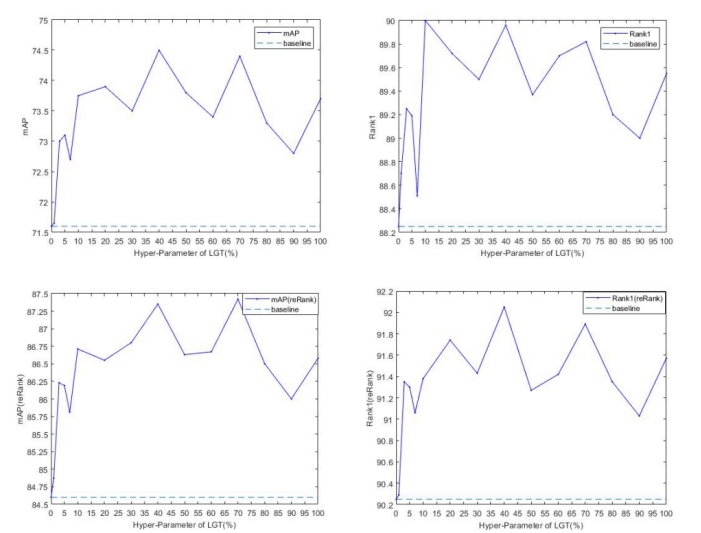}
	\caption{Performance of LGT under different hyperparameters on Market1501.}
\end{figure}
\subsection{Hyper-Parameter Setting}
During CNN training, two hyper-parameters need to be evaluated. One of them is GGT probability $p$. Firstly, we take the hyper-parameter $p$ as 0.01, 0.03, 0.05, 0.07, 0.1, 0.2, 0.3,..., 1 for the GGT experiments. Then we take the value of each parameter for three independent repetitions of the experiments. Finally, we calculate the average of the final result. The results of different p are shown in Fig.$\color{red}3$. We can see that when $p=0.05$, the performance of the model reaches the maximum value in Rank-1 and mAP. If we do not specify, the hyper- parameter is set $p=0.05$ in the next experiments.

Another hyper-parameter is LGT probability $p_r$. We take the hyper-parameter $p_r$ as the same as above for the LGT experiments, whose selection process is similar to the above $p$. The results of different $p_r$ are shown in Fig.$\color{red}4$.

Obviously, when $p_r=0.4$ or $p_r=0.7$, the model achieves better performance. And the best performance is achieved when $p_r=0.4$. If we do not specify, the hyper- parameter is set $p_r=0.4$ in the subsequent experiments.


\begin{figure}[t]
	\setlength{\abovecaptionskip}{0.1cm}
	\setlength{\belowcaptionskip}{-0.4cm}   
	\centering
	\includegraphics[width=1\linewidth]{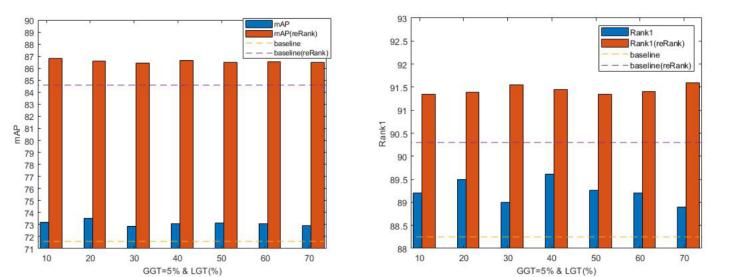}
	\caption{Performance of combining GGT with LGT under different hyperparameters on Market1501.}
\end{figure}

\textbf{Evaluation of GGT and LGT}. Compared with the best results of GGT on baseline\cite{zheng2018discriminatively}, the accuracy of LGT is improved by 0.5\% and 1.4\% on Rank-1 and mAP, respectively. Under the same conditions using re-Ranking\cite{re_ranking}, the accuracy of Rank-1 and mAP is improved by 0\% and 0.4\%, respectively. Therefore, the advantages of LGT are more obvious when re-Ranking is not used. However, Fig.$\color{red}4$ also shows that the performance improvement brought by LGT is not stable enough because of the obvious fluctuation in LGT, while the performance improvement brought by GGT is very stable. Therefore, we improve the stability of the method by combining GGT with LGT.

\textbf{Evaluation by Combining GGT with LGT}. First, we fix the hyper-parameter value of GGT to $p=0.05$, then keep the control variable unchanged to further determine the hyper-parameter of LGT. Finally, we take the hyper-parameter pr of LGT to be 0.1, 0.2, ···, 0.7 to conduct combination experiments of GGT and LGT, and conduct 3 independent repeated experiments for each parameter $p_r$ to get the average value. The result is shown in $\color{red}5$. It can be seen that the performance improvement brought by the combination of GGT and LGT is more stable and with less fluctuation, and the comprehensive performance of the model is the best when the hyper-parameter value of LGT is $p_r=0.4$.

\begin{figure}[t]
	\setlength{\abovecaptionskip}{0.1cm}
	\setlength{\belowcaptionskip}{-0.4cm}   
	\centering
	\includegraphics[width=1\linewidth]{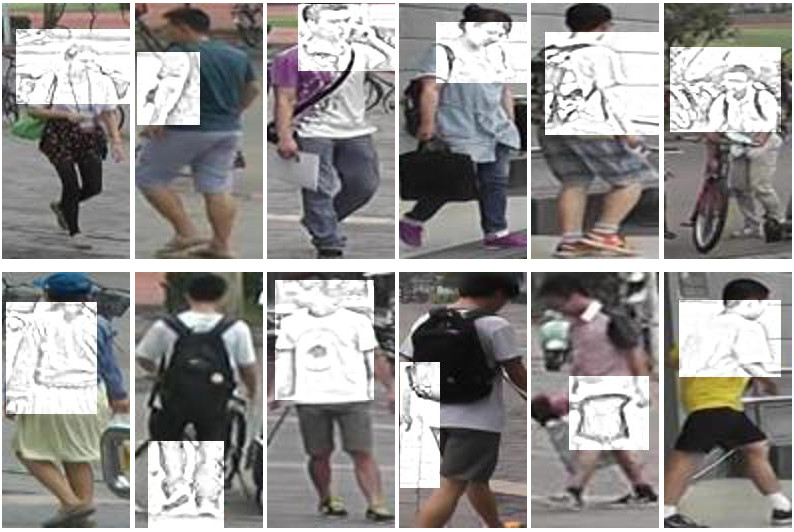}
	\caption{Diagram of Global Sketch Transformation (GST) and Local Sketch Transformation (LST).}
\end{figure}

\subsection{Comparison Experiments}

\textbf{Performance comparison and analysis}. We first evaluate  baseline \cite{zheng2018discriminatively} on the Market-1501 dataset\cite{market1501}. To be consistent with recent works, we follow the new training/testing protocol to conduct our experiments by k-reciprocal re-ranking (RK)~\cite{re_ranking}. As can be seen from Fig.$\color{red}3$ and  Fig.$\color{red}4$, our method improves by 1.2\% on Rank-1 and 3.3\% on mAP on the baseline, and 1.5\% on Rank-1 and 2.1\% on mAP above baseline in the same conditions using the re-Ranking\cite{re_ranking}.

Secondly, we further test the method in this paper on the baselines\cite{stong_baseline,FastReID} with better performance. As we can see from Table.$\color{red}1$ to Table.$\color{red}3$, the best results of our method improve by 0.6\% and 1.3\% on the Rank-1 and mAP on the strong baseline\cite{stong_baseline}, respectively, and 0.8\% and 0.5\% Rank-1 and mAP above baseline under the same conditions using the re-Ranking\cite{re_ranking}, respectively. On fastReID\cite{FastReID}, our method is 0.2\% higher and 0.9\% than baseline in Rank-1 and mAP, respectively, and higher 0.1\% and 0.3\% than baseline under using re-Ranking. 

The default configuration on the Strong Baseline\cite{stong_baseline} and FastReID\cite{FastReID} uses data augmentation such as random flipping\cite{simonyan2014very}, cropping\cite{krizhevsky2012imagenet}, and erasing\cite{zhong2020random}. The method proposed in this paper further improves the model accuracy on the basis of using them, which shows that our method can be combined with other data augmentation methods.

\begin{table}[]
	\centering 
	\setlength\tabcolsep{3pt}
	\caption{Performance comparison on Market1501 dataset.}
	\begin{tabular}{ccc}
		\hline
		\multirow{2}{*}{Methods}         & \multicolumn{2}{c}{Market1501}            \\ \cline{2-3} 
		& Rank-1(\%)          & mAP(\%)             \\ \hline
		IANet\cite{IANet}             & 94.4                & 83.1                \\ 
		DGNet\cite{zheng2019joint}           & 94.8                & 86.0                \\ 
		SCAL\cite{SCAL}             & 95.8                & 89.3                \\ 
		Circle Loss\cite{Circle}    & 96.1                & 87.4                \\ 
		SB\cite{stong_baseline} & 94.5                & 85.9                \\ \hline
		SB\cite{stong_baseline} + RK\cite{re_ranking}                & 95.4                & 94.2                \\ 
		SB + GGT(ours)             & \textbf{94.6} & 85.7                \\ 
		SB + GGT+ RK(ours)     & \textbf{96.2} & \textbf{94.7} \\ 
		SB + LGT(ours)             & \textbf{95.1} & \textbf{87.2} \\ 
		SB + LGT + RK(ours)    & \textbf{95.9} & \textbf{94.4} \\ \hline
		FastReID\cite{FastReID}       & 96. 3               & 90.3                \\ 
		FastReID + RK                & 96.8                & 95.3                \\ 
		FastReID + GGT(ours)             & \textbf{96.5} & \textbf{91.2} \\ 
		FastReID + GGT + RK(ours)    & \textbf{96.9} & \textbf{95.6} \\ \hline
	\end{tabular}
\end{table}

\begin{table}[]
	\centering 
	\setlength\tabcolsep{3pt}
	\caption{Performance comparison on DukeMTMC dataset.}
	\begin{tabular}{ccc}
		\hline
		\multirow{2}{*}{Methods}         & \multicolumn{2}{c}{DukeMTMC}              \\ \cline{2-3} 
		& Rank-1(\%)          & mAP(\%)             \\ \hline
		IANet\cite{IANet}           & 87.1                & 73.4                \\ 
		DGNet\cite{zheng2019joint}            & 86.6                & 74.8                \\ 
		SCAL\cite{SCAL}         & 89.0                & 79.6                \\ \hline
		SB\cite{stong_baseline} & 86.4                & 76.4                \\ 
		SB + RK\cite{re_ranking}                 & 90.3                & 89.1                \\ 
		SB + GGT(ours)             & \textbf{87.8} & \textbf{77.3} \\ 
		SB + GGT+ RK(ours)     & \textbf{90.9} & \textbf{89.2} \\ 
		SB + LGT(ours)             & \textbf{87.3} & \textbf{77.3} \\ 
		SB + LGT + RK(ours)    & \textbf{91}   & \textbf{89.4} \\ \hline
		FastReID\cite{FastReID}      & {92.4}       & {83.2}       \\ 
		FastReID + RK                & 94.4                & 92.2                \\ 
		FastReID + LGT(ours)             & \textbf{92.8} & \textbf{84.2}   \\ 
		FastReID + LGT + RK(ours)    & \textbf{94.3}       & \textbf{92.7} \\ \hline
	\end{tabular}
\end{table}

\begin{table}[]
	\centering 
	\setlength\tabcolsep{3pt}
	\caption{Performance comparison on MSMT17 dataset.}
	\begin{tabular}{ccc}
		\hline
		\multirow{2}{*}{Methods}      & \multicolumn{2}{c}{MSMT17}                \\ \cline{2-3} 
		& Rank-1(\%)          & mAP(\%)             \\ \hline
		IANet\cite{IANet}       & 75.5                & 46.8                \\ 
		DGNet\cite{zheng2019joint}         & 77.2                & 52.3                \\ 
		RGA-SC\cite{zhang2020relation}     & 80.3                & 57.5                \\ 
		SCSN\cite{chen2020salience}        & 83.8                & 58.5                \\ 
		AdaptiveReID\cite{ni2021adaptive} & 81.7                & 62.2                \\ \hline
		FastReID\cite{FastReID}             & 85.1                & 63.3                \\ 
		FastReID + GGT(ours)          & \textbf{86.2} & \textbf{65.3}   \\ 
		FastReID + GGT\&LGT(ours)     & \textbf{86.2} & \textbf{65.9} \\ \hline
	\end{tabular}
\end{table}

\textbf{Cross-domain tests}. Cross-domain person re-identification aims at adapting the model trained on a labeled source domain dataset to another target domain dataset without any annotation. It is pointed out by\cite{stong_baseline} that the higher accuracy of the model does not mean that it has better generalization capacity. In response to the above potential problems, we use cross-domain tests to verify the robustness of the model. Experiments show that the proposed method effectively enhances the generalization capacity of the model, and the Table $\color{red}2$ shows the cross-domain experiments of the proposed method between two datasets, Market-1501\cite{market1501} and DukeMTMC\cite{duke}. In order to further explore the  effectiveness  of the proposed method in cross-domain experiments, we use GGT to conduct the following cross-domain experiments on strong baseline\cite{stong_baseline}.  The   experiments  are  shown  in Table.$\color{red}4$. 

In Table 4, +REA means that the trick of Random Erasing is used in model training, -REA means turning it off. Experimental results show that random  erasing\cite{zhong2020random} can also significantly improve the performance of the ReID model, but it will cause a significant drop in cross-domain performance. The proposed method can not only  significantly improve the cross-domain performance of the  ReID model, but also be more robust because of learning more discriminative features.

\begin{table}[]\small
	\centering 
	\setlength\tabcolsep{1pt}
	\caption{The performance of different models is evaluated on cross-domain dataset. M→D means that we train the model on Market1501\cite{market1501} and evaluate it on DukeMTMC\cite{duke}.}
	\begin{tabular}{cclcl}
		\hline
		\multirow{3}{*}{Methods}       & \multicolumn{4}{c}{Cross-Domain}                                                \\ \cline{2-5} 
		& \multicolumn{2}{c}{M→D}  & \multicolumn{2}{c}{D→M}  \\ \cline{2-5} 
		& \multicolumn{2}{c}{Rank-1/mAP(\%)}     & \multicolumn{2}{c}{Rank-1/mAP(\%)}     \\ \hline
		SB\cite{stong_baseline}+REA\cite{zhong2020random}+RK            & \multicolumn{2}{c}{33.6/24.3}          & \multicolumn{2}{c}{51.6/32.3}          \\ 
		SB+REA+GGT+RK(ours) & \multicolumn{2}{c}{\textbf{37.8/27.8}} & \multicolumn{2}{c}{\textbf{55.4/35.7}} \\ 
		SB-REA+RK            & \multicolumn{2}{c}{45.5/37.0}          & \multicolumn{2}{c}{58.2/37.8}          \\ 
		SB-REA+GGT+RK(ours) & \multicolumn{2}{c}{\textbf{48.2/37.9}} & \multicolumn{2}{c}{\textbf{65.0/43.7}} \\ \hline
	\end{tabular}
\end{table}

\begin{table}[]
	\centering 
	\setlength\tabcolsep{8pt}
	\caption{Performance comparison between our LGT conversion and DGNet data augmentation on Market1501.}
	\begin{tabular}{cclcl}
		\hline
		\multirow{2}{*}{Methods} & \multicolumn{4}{c}{Market1501}                                        \\ \cline{2-5} 
		& \multicolumn{2}{c}{Rank-1}        & \multicolumn{2}{c}{mAP(\%)}       \\ \hline
		Baseline\cite{zheng2018discriminatively}            & \multicolumn{2}{c}{88.8}          & \multicolumn{2}{c}{71.6}          \\ 
		Baseline + DGNet\cite{zheng2019joint}       & \multicolumn{2}{c}{{88.9}} & \multicolumn{2}{c}{{72.1}} \\ 
		Baseline+ LGT(ours) & \multicolumn{2}{c}{\textbf{90.0}} & \multicolumn{2}{c}{\textbf{74.9}} \\ \hline
	\end{tabular}
\end{table}

\begin{table}[]
	\centering 
	\setlength\tabcolsep{3pt}
	\caption{Cross-domain performance comparison between our LGT and DGNet on Market1501.}
	\begin{tabular}{cclcl}
		\hline
		\multirow{2}{*}{Methods} & \multicolumn{4}{c}{Market1501→DukeMTMC}                               \\ \cline{2-5} 
		& \multicolumn{2}{c}{Rank-1}        & \multicolumn{2}{c}{mAP}           \\ \hline
		Baseline\cite{zheng2018discriminatively}            & \multicolumn{2}{c}{37.8}          & \multicolumn{2}{c}{27.0}          \\ 
		Baseline + DGNet\cite{zheng2019joint}       & \multicolumn{2}{c}{{36.7}} & \multicolumn{2}{c}{{25.6}} \\ 
		Baseline+ LGT(ours) & \multicolumn{2}{c}{\textbf{39.7}} & \multicolumn{2}{c}{\textbf{27.9}} \\ \hline
	\end{tabular}
\end{table}

\begin{table}[]
	\centering 
	\setlength\tabcolsep{3pt}
	\caption{Performance comparison on Market1501 dataset.}
	\begin{tabular}{cclcl}
		\hline
		\multirow{2}{*}{Methods}                            & \multicolumn{4}{c}{Market1501}                                        \\ \cline{2-5} 
		& \multicolumn{2}{c}{Rank-1(\%)}    & \multicolumn{2}{c}{mAP(\%)}       \\ \hline
		Baseline\cite{zheng2018discriminatively}                                       & \multicolumn{2}{c}{88.8}          & \multicolumn{2}{c}{71.6}          \\ 
		Baseline + RK\cite{re_ranking}                                  & \multicolumn{2}{c}{\textbf{90.5}} & \multicolumn{2}{c}{\textbf{85.2}} \\ 
		Baseline+ GST+LST(ours)                        & \multicolumn{2}{c}{\textbf{88.9}} & \multicolumn{2}{c}{\textbf{72.6}} \\ 
		\multicolumn{1}{l}{Baseline+ GST+LST+RK(ours)} & \multicolumn{2}{c}{\textbf{91.2}} & \multicolumn{2}{c}{\textbf{86.8}}  \\ \hline
	\end{tabular}
\end{table}

\begin{table}[]\small
	\centering 
	\setlength\tabcolsep{2pt}
	\caption{Performance comparison between our RCD and Adversarial Feature Learning on Sketch ReID dataset.}
	\begin{tabular}{cccc}
		\hline
		Sketch ReID dataset          & Rnak-1(\%) & Rnak-5(\%) & Rank-10(\%) \\ \hline
		AFL\cite{sketch-criminal} & 34.0       & 56.3       & 72.5        \\ 
		GST+LST(ours)                & \textbf{42.5}      & \textbf{70.0}       & \textbf{87.5}        \\ \hline
	\end{tabular}
\end{table}

\begin{figure}[]
	\setlength{\abovecaptionskip}{0.1cm}
	\setlength{\belowcaptionskip}{-0.4cm}   
	\centering
	\includegraphics[width=0.8\linewidth]{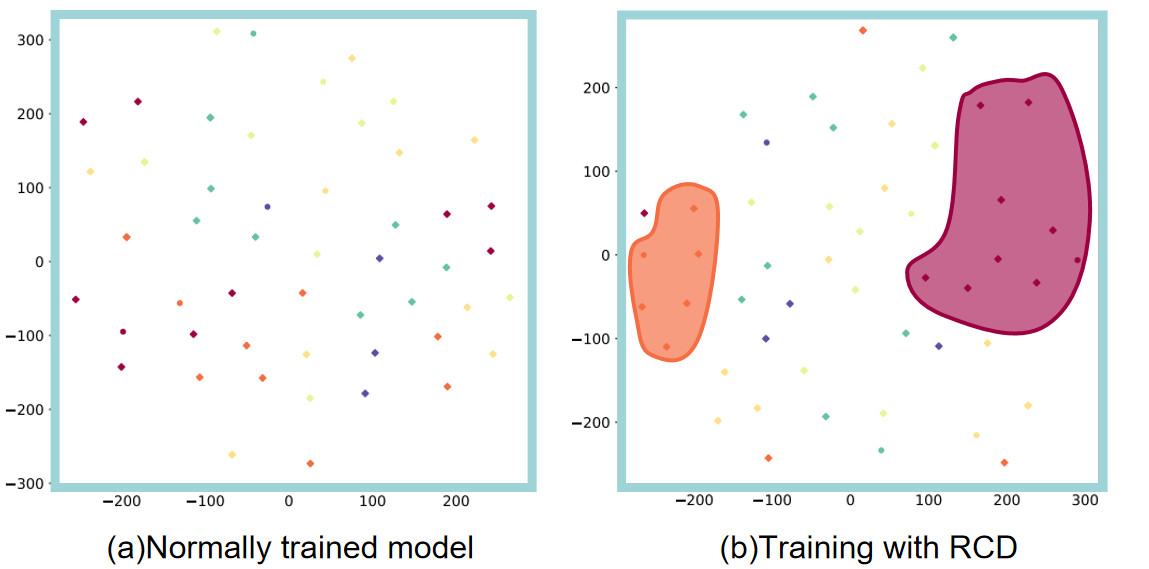}
	\caption{t-SNE~\cite{tsne} visualization of six randomly selected images with different identities on Market1501\cite{market1501}. Each image corresponds to the randomly generated images with color deviation. The same color means that they are obtained by transformation of the same image. Dots means original example.}
\end{figure}

\textbf{Comparison of state-of-the-arts}. A comparison of the performance of our method with the state-of-the-art methods DGNet\cite{zheng2019joint} on Market1501\cite{market1501} is shown in Table.$\color{red}5$ and Table.$\color{red}6$. As can be seen from Table.$\color{red}5$, our method delivers a performance improvement that far exceeds that of DGNet, the state-of-the-art GAN-based method, by more than 2.7 percentage points on mAP, which suggests that the proposed method is superior to existing data augmentation. 

As can be seen from Table.$\color{red}6$, the generalization ability of the proposed method in cross-domain tests is improved by 1.9 percentage points in the Rank-1 compared with the baseline\cite{zheng2018discriminatively}, which further shows that the proposed method is better than the existing data augmentation based on generative models. It is worth noting that when the data generated by DGNet is used for model training, the cross-domain performance of the model is poorly, which confirms the point of this paper that color deviation is is difficult to exhaust and that instead of enhancing robustness to input changes by generating a variety of data for the model to "see" during training, it is better to implicitly reduce the weight of the model in the discriminant feature of color information.

\textbf{Cross-modal retrival}. Another form of strategy proposed in this paper is to take sketch image as the intermediary of balancing weight. By applying the proposed global homogeneity transformation and local homogeneity transformation, the sketch image is transformed as a homogeneous image, as shown in Figure.$\color{red}6$. It can not only improve the robustness of the model, but also realize the sketch-based ReID. We can see this clearly from Table.$\color{red}.7$ and Table.$\color{red}.8$.

In terms of cross-modal retrieval\cite{song2017deep,sketch-criminal,basaran2020efficient}, in order to match images of different modalities, existing approaches usually achieve cross-modal retrieval with the help of attention mechanisms\cite{song2017deep}, multi-stream networks\cite{basaran2020efficient}, and generative adversarial networks\cite{sketch-criminal}. Lu et al. proposed a cross-domain adversarial feature learning (AFL) method for sketch re-identification and contributed the sketch character re-identification dataset\cite{sketch-criminal}.

In order to make a fair comparison, as same as AFL\cite{sketch-criminal}, the method proposed in this paper is firstly trained on the Market-1501 dataset, and then fine tuned on sketch ReID dataset. In parameter setting, this paper set 5\% Global Sketch Transformation and 70\% Local Sketch Transformation. The experiment result shows that the performance improvement in the Sketch Re-identification more than 8\%. This experiment also shows the generality of the proposed method.

\begin{figure}[]
	\setlength{\abovecaptionskip}{0.1cm}
	\setlength{\belowcaptionskip}{-0.4cm}   
	\centering
	\includegraphics[width=1\linewidth]{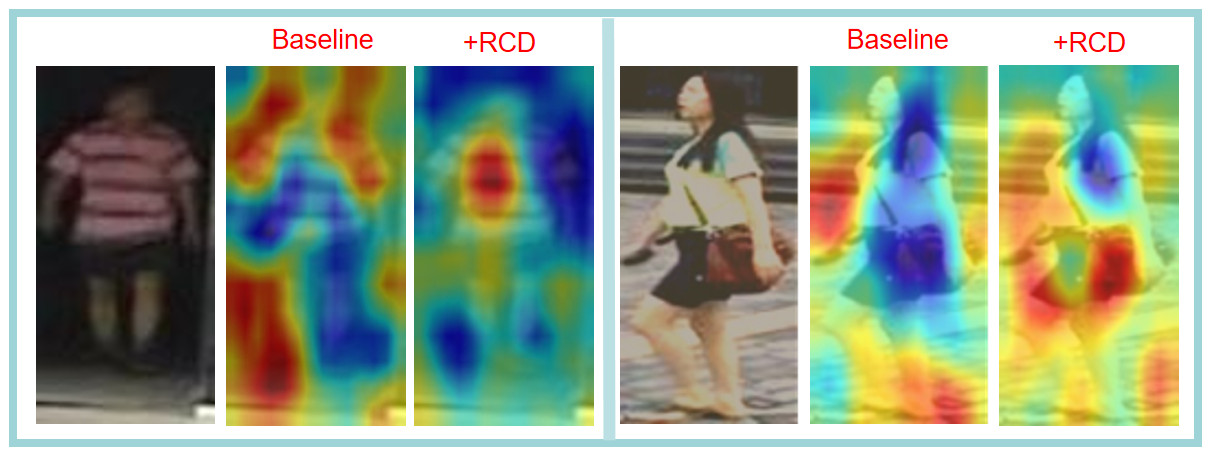}
	\caption{Comparison of Grad-CAM~\cite{Grad-CAM} activation map between normally trained model and our proactive defense model.}
\end{figure}

\textbf{Visualization analysis}. As the show in Figure $\color{red}7$, the model trained by DCR is robust to color variations. Therefore, we can observe that the features of examples with color deviation exhibit clustering effects better.

Grad-CAM~\cite{Grad-CAM} uses the gradient information flowing into the last convolutional layer of the CNN to visualize the importance of each neuron in the output layer for the final prediction, by which it is possible to visualize which regions of the image have a significant impact on the prediction of a model. As shown in Figure $\color{red}8$, we can see that the the normally trained model activates irrelevant parts in the case of severe color deviation, while the model trianed with RCD is still effectively activating some important parts.


\section{Conclusion}
In this paper, a simple, effective and general strategy that can be applied in computer vision to overcome color deviation. Neither does the method require large scale training like GAN, nor introduces any noise.  The method uses a random homogeneous transformation to realize the modeling of different modal relationships. The model balances the weights between color features and discriminative non-color features by fitting differentiated homogeneous information in a mixed  domain with information bias during the training process, thus reducing the negative impact of color deviation on ReID. In addition, this paper reveals the intrinsic reasons why networks trained with RCD outperform ordinary networks from a classification perspective. At the same time, experiments on several datasets and baselines show that the proposed method is effective and outperforms the state of the arts, and extra experiments show that the proposed strategy has natural cross modal properties.

{\small
\bibliographystyle{unsrt}
\bibliography{egbib}
}

\end{document}